%% file: report.tex
\documentclass[12pt, a4paper,twoside, times, openany]{book}
\usepackage{fullpage}
\usepackage{setspace}

\usepackage{geometry}
\usepackage{graphicx}
\graphicspath{{figures/}}
\usepackage{epstopdf}
\usepackage{amsmath}
\usepackage{amsthm}
\usepackage{amssymb}
\usepackage{booktabs}
\usepackage{multirow}
\usepackage{bm} 
\usepackage[hidelinks]{hyperref}
\usepackage{subcaption} 
\usepackage{siunitx}
\usepackage{tocloft}

\usepackage{color,soul,framed}   
\usepackage{xcolor,soul,framed}
\colorlet{shadecolor}{yellow}

\makeatletter
\newcommand\listoftablesandfigures{%
	\chapter*{List of Tables and Figures}%
	\phantomsection
	\@starttoc{lot}%
	\bigskip
	\@starttoc{lof}}
\makeatother

\usepackage{fancyhdr}
\usepackage{titlesec}
\titleformat{\chapter}[display]{\bfseries\boldmath\huge}%
{\chaptername\ \thechapter}{1em}{}{}
\titlespacing{\chapter}{0pt}{-8.0ex}{10\wordsep}

\fancypagestyle{plain}{
	\fancyhead{} 
	\fancyhead[RE]{}
	\fancyhead[LO]{}
	\fancyfoot[C]{\thepage}

}

\begin{document}
\frontmatter
\newgeometry{top=2.5cm,bottom=2.5cm,outer=2.5cm,inner=2.5cm} 
\include{./chapters/cover_page}

\setlength{\parindent}{20pt}

\include{./chapters/preface_page}

\vspace{-2.2cm}
\tableofcontents


\renewcommand{\cleardoublepage}{\newpage}
\mainmatter
\setlength{\parindent}{20pt}
\include{./chapters/ch1_background}
\include{./chapters/ch2_research_goal}
\include{./chapters/ch3_methods}

\include{./chapters/ch4_results}
\include{./chapters/ch5_conclusion}

\begingroup
\setstretch{1}
\bibliographystyle{./references/IEEEtran}
\bibliography{./references/demo}
\endgroup

\end{document}

%% file: chapters/cover_page.tex
\newpage\thispagestyle{empty}


		
		
		
		
		
		
		
		
			
	
	




\hspace*{3cm}
\begin{center}
    {\bf {\Large Diffusion idea exploration \\[1.5mm] for art generation}

		\vfill
		
		by
		
		Nikhil Verma
		
		\vfill
		
		while working under the guidance of 
		
		\begin{figure}[tbh]
			\centerline{\includegraphics[width=1.5in]{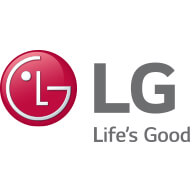}}
		\end{figure}
		Industrial Supervisor: Kevin Ferreira 
		
		Senior Director, LG Toronto AI Lab
		
		\vfill
		
		\begin{figure}[tbh]
			\centerline{\includegraphics[width=1.5in]{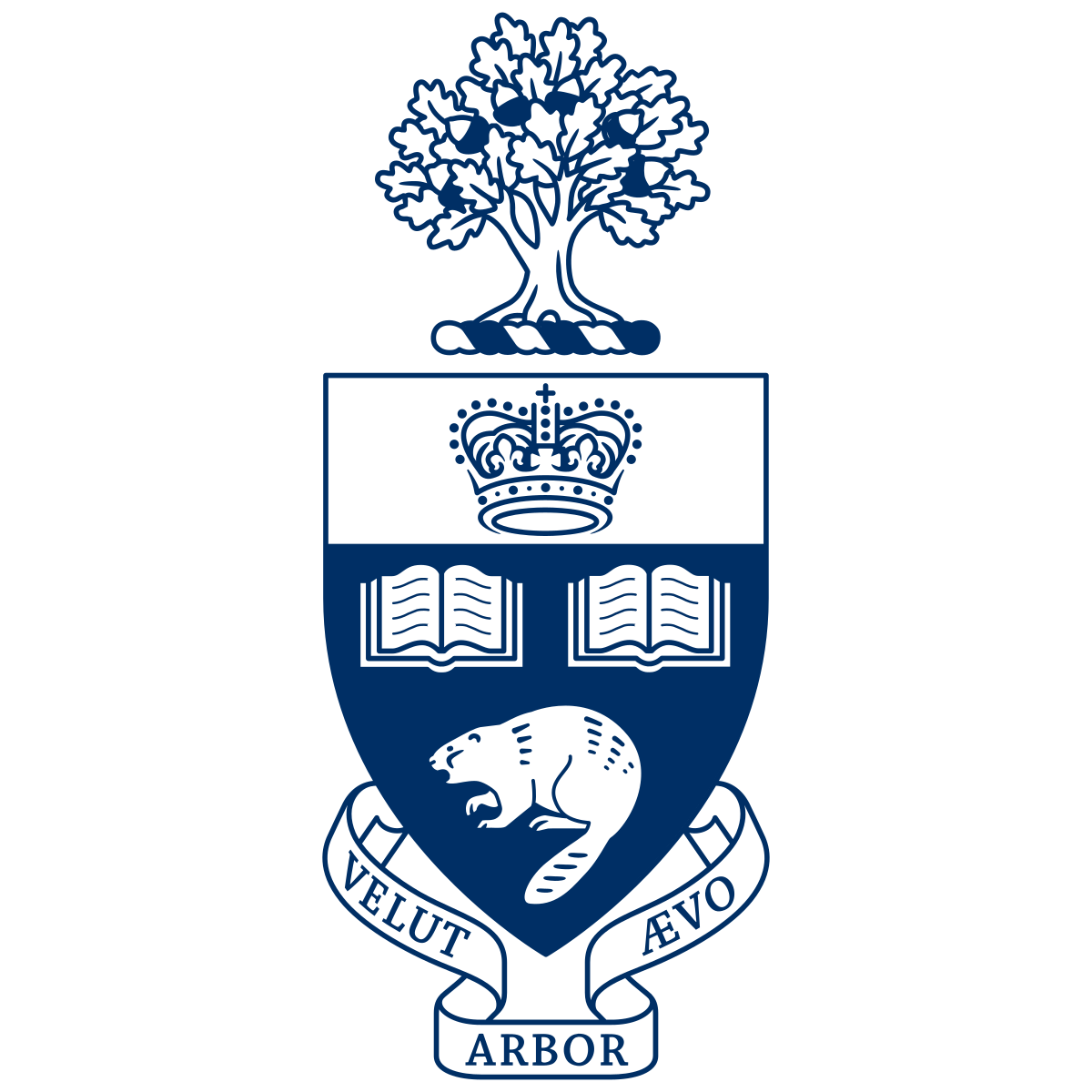}}
		\end{figure}
		
		Academic Supervisor: Dr. Scott Sanner 
		
		Department of Mechanical and Industrial Engineering

		
		
	    \bf{\Large Master of Science in Applied Computing }
		
		2022
		
		\vfill
		\begin{center}
			Department of Computer Science
			
			University of Toronto \end{center}
	}
	\vfill
	
\end{center}


\newpage\pagestyle{plain}
\thispagestyle{empty}
\noindent
\bigskip

\vspace{6.0cm}
\begin{center}
	\text{Dedicated to my parents}
\end{center}


%% file: chapters/preface_page.tex


\newpage\thispagestyle{plain}

\addcontentsline{toc}{chapter}{Acknowledgments}

\begin{center}
	{\bf \large Acknowledgments}
\end{center}

Words will not be sufficient to thank my supervisors, Dr. Scott Sanner and Kevin Ferreira whose constant guidance and support that motivated me during the 8 months long internship for the degree requirement of Master of Science in Applied Computing at University of Toronto, Toronto, Canada. Without their assistance and involvement in each and every step of the process, this internship would not have been successful. I would like to thank them for their understanding over the internship tenure. Thanks to the organisation LG Electronics Canada for providing me an opportunity to work on interesting problem and also University of Toronto for equipping me academically and giving me an opportunity to make connection with such a great organisation for research internship. Multiple brain storming sessions with Ehsan Nezhadarya, Vic Huang, Chao Wang and Manasa Bharadwaj helped to think of various usecases of utilising diffusion architecture for industrial applications.

I extend my word of thanks to my colleagues for their reading and editing help which served to give perfect shape to this report document. The seminar session discussions helped in logic building.

A big word of thanks to the Mitacs Organisation to fund my internship and for providing support for materials and research cost involved. Its only because of them that I was able to focus on my work instead of thinking about exploring other ways to finance my study and internship tenure. 

This journey would not have been possible without the support and mentorship from Krishna Prasad, Dr. Pradeep Kumar, Dr. Prashant singh Rana, Dr. Neeraj Kumar, Dr. Geeta Kasana and Pooja Dubey. I express my gratitude to Parmatma(the divine soul) for blessing me with beautiful souls in form of my parents, sibling and friends who provided encouragement through phone calls and messages despite my own devotion to correspondence.   


\newpage\thispagestyle{plain}

\addcontentsline{toc}{chapter}{Abstract}

\begin{center}
	{\bf Abstract}
\end{center}

Cross-Modal learning tasks have picked up pace in recent times. With plethora of applications in diverse areas, generation of novel content using multiple modalities of data has remained a challenging problem. To address the same, various generative modelling techniques have been proposed for specific tasks. Novel and creative image generation is one important aspect for industrial application which could help as an arm for novel content generation. Techniques proposed previously used Generative Adversarial Network(GAN), autoregressive models and Variational Autoencoders (VAE) for accomplishing similar tasks. These approaches are limited in their capability to produce images guided by either text instructions or rough sketch images decreasing the overall performance of image generator. We used state of the art diffusion models to generate creative art by primarily leveraging text with additional support of rough sketches. Diffusion starts with a pattern of random dots and slowly converts that pattern into a design image using the guiding information fed into the model. Diffusion models have recently outperformed other generative models in image generation tasks using cross modal data as guiding information. The initial experiments for this task of novel image generation demonstrated promising qualitative results.






%% file: chapters/ch1_background.tex
\chapter{Background Information}
\label{ch1_intro} 

\section{Company description}
The company LG Electronics(LGE) is a global brand in the market of technology and consumer electronics. It has its presence in almost each and every country on the planet and provides employment to more than 75000 individuals across the globe. Major part of global sales of LG comes from its four arms namely 
\begin{enumerate}
    \item Home Appliance \& Air Solution
    \item Home Entertainment
    \item Vehicle component Solutions and 
    \item Business Solutions
\end{enumerate}

LGE is a leading manufacturer of many consumer products such as TVs, home appliances, monitors, service robots, air solutions and automotive components. Its premium products are famous under the brand name of LG Signature while LG ThinQ provide intelligent products. LG ThinQ products have AI technology that evolves functionality by analyzing and learning your lifestyle, habits and preferences.

\section{Problem statement}
\begin{quote}
    \center ``A Good Sketch is Better Than a Long Speech"
\end{quote}
The notion of importance of images over text as mentioned in the famous quote of Napoleon Bonaparte has been stressed by various leaders, philosophers and thinkers of all times. So, Text-to-Image(T2I) generation has become an attractive area of work for the deep learning research community where the generated image is a depiction of a components mentioned using text modality.

T2I methods have provided a pipeline for transitioning from one modality to another. Although text could be used to narrate descriptive attributes of images such as colors, objects and styles, other non descriptive attributes such as pose, structure and relative arrangement could be best described using a guide image. Text and Image guided design generation, which is a sister branch of T2I, uses both text and images to generate modified images where the control of generation is shared between more than one modality of input. Multiple modalities help in providing both the content and context for generating data. 

The goal of this project is to generate creative artistic images using textual or visual guidance. Generating images is crucial for developing new ideas for creating products for industrial applications. Main benefit of generative art is to frequently create novel design patterns inspired from distribution of old existing patterns. The output of generative models represent designer's high level artistic vision.

\begin{figure}[htb]
  \centering
   \centerline{\includegraphics[width=1\linewidth]{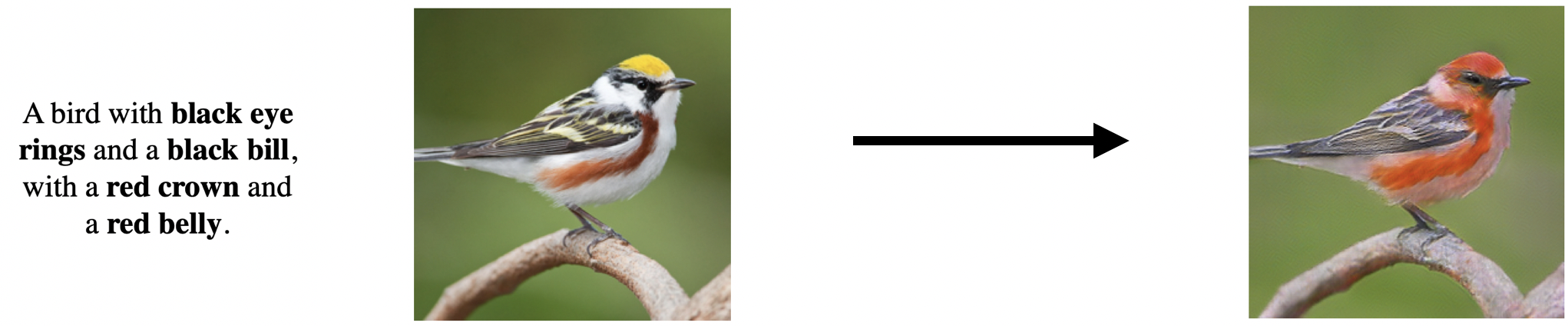}}

   \caption{Manipulation of reference image using text \cite{li2020manigan}}
   \label{fig:imageManipulation}
\end{figure}

The task of text and image guided design generation was inspired closely from Text based image manipulation, as proposed by the authors of \cite{li2020manigan}, where focus is on learning to manipulate the base image using text instruction. An illustration where the image of a bird served with modification text having “black eye rings, red crown and red belly” should produce another accordingly modified image is shown in Fig \ref{fig:imageManipulation}.

With many industrial appliances directly interacting with the customers everyday, it is in the natural favor of industries to develop their arm for working on AI Research trends that can be used in improvising the customer experience of using appliances and products. Working on research and development of text and image based art generation task will help the designers who are constantly involved in thinking of design variations of the range of appliances. This will help designers by providing abundance of initial designs to brainstorm and will save their time to create such initial drafts. 

\section{Contributions}

In this work, we used \textsc{Stable diffusion} \cite{rombach2022high} model. It is a diffusion based model in the latent space where the  diffusion happens in the shell of a VAE. It helps in generating novel image and text based artistic design patterns. Stable diffusion model uses a U-net architecture augmented with classifier guidance using the conditioning variable. Model was learned in a classifier-free guidance style which trains both conditioned and unconditioned model to find direction for early convergence and provide more control in the generation process. Inspired by GLIDE\cite{nichol2021glide}, the conditional information uses both text and image embeddings to guide the reverse diffusion process. For analysing the proposed pipeline, we used manually created dataset of simple design images and textual prompts to guide the diffusion learning.

%% file: chapters/ch2_research_goal.tex
\chapter{Research Goals and Outcomes}
\label{ch2_research_goal} 

\section{Goal of the project}
Most of the T2I image generation methods have focused on adversarial learning using generative adversarial network(GAN) \cite{goodfellow2020generative} approaches or likelihood based latent methods using auto-regressive transformer(ART) combined with variational auto-encoder(VAE) \cite{kingma2013auto}. GANs suffer from mode-collapse, training instability and limited inversion performance while likelihood based models use billions of parameters for learning distributions and are therefore computationally expensive with non-parallelizable sequential generation processes. Another class of promising generative method called Diffusion probabilistic model beats GAN for image synthesis \cite{dhariwal2021diffusion} and produces astonishing results using text guidance. In creative art generation task we wish to utilise the high level direction provided by designer of the art to generate final art designs. We used Stable diffusion for accomplishing this task. The initial direction could be in form of multiple modalities of input such as text describing expected design or rough sketch image of desired art. The goal of this project is to use text and image guided denoising diffusion probabilistic model for novel art design generation.

\section{Approach and contributions}
The modelling approach used for solving this task was based on Diffusion. Diffusion model uses forward and reverse diffusion processes. Forward diffusion is a Markovian chain that adds Gaussian noise at each step of the process to an image data point until it appears to come from standard normal distribution, making the image noisier at each step. Reverse diffusion learns to convert noisy images back to an original image by estimating likelihood of the added noise, denoising the image at each step.

In summary my contribution through this work is two fold:-
\begin{enumerate}
    \item To clearly understand and use latent diffusion based artistic designer image generator which uses fused text and image embedding as conditioning parameters to control the flow of the generation process.
    
    \item To conduct experiments on manually generated dataset for the process of creative art design using generative model. The results of experiments show that diffusion based generation produce interesting results that could be further used  by human art designer in their brainstorming sessions.  
\end{enumerate}

\section{Related Work}
\textbf{Text to Image generation}: Starting with GAN \cite{reed2016generative} \cite{zhang2017stackgan} \cite{zhang2018stackgan++} \cite{xu2018attngan} \cite{li2019controllable} \cite{qiao2019mirrorgan} based architectures which evolved the idea of guiding the image generation through text, earlier work used text matching through discriminator and data augmentation techniques to accommodate image generation with varied text inputs. Despite the novel contributions, each approach was an advancement of GAN based architecture. Working on likelihood based modeling(autoregressive modeling of discrete image representation) and aiming to learn cross-modal data distribution, DALL-E \cite{ramesh2021zero} proposed a transformer-like autoregressive approach which used quantized image tokens generated from discrete-VAE to represent images in latent space. The transformer-decoder then learned the association between text tokens and discrete image tokens by maximizing evidence lower bound of the joint likelihood of the cross-modal distribution. 

DDPM\cite{ho2020denoising} proposed another generative technique for Image synthesis based on diffusion modeling. Later ADM\cite{dhariwal2021diffusion} proved the capability of diffusion models to beat GAN on image quality metrics. It also introduced the idea of conditioning the diffusion model on any guiding information such as labels to control the image generation process. GLIDE\cite{nichol2021glide} used this idea of conditioning image generation on text caption to generate images out of text. CogView\cite{ding2021cogview} independently proposed the same idea as DALL-E but released later than that and uses stable training techniques like PB relax and sandwich layer normalization. Make-A-Scene\cite{gafni2022make} used text and optional segmentation map of the scene to be generated in a autoregressive fashion of discrete image representation generation. DALL-E 2\cite{ramesh2022hierarchical}, LAFITE\cite{zhou2022towards}, IMAGEN\cite{saharia2022photorealistic}, PARTI \cite{yu2022scaling}, Stable diffusion \cite{rombach2022high} are some other latest works that build larger parametric models for more realistic image generation based on text modality.

\textbf{Image Manipulation using text}: Unlike text-to-image generation with more flexibility in the generation process, text guided image manipulation aims to semantically edit only the parts/attributes of the image mentioned in the text. ManiGAN \cite{li2020manigan} instead of joining text information along the channel  direction, uses two components to manipulate images namely Affine Combination Module and Detailed Correction Module. DiffusionCLIP \cite{kim2022diffusionclip} performs faithful image manipulation leveraging diffusion modeling for manipulated image generation. Text-as-neural-operator \cite{zhang2021text} focuses on synthetic image manipulation using GAN approach for not only changing some descriptive attribute information but performing some actionable manipulation of images such as adding, modifying and removing objects from a scene. All these methods take one initial image to be manipulated and a single text instruction describing the manipulation demanded.

Building on top of one time image manipulator, in multi-turn manipulator we have a sequence of text instructions to manipulate an image iteratively. GeNeVA \cite{el2019tell} is the pioneering work in this domain which proposes two datasets namely CoDraw \cite{kim2017codraw} and iClevr \cite{johnson2017clevr}. To deal with the problem of scarcity of data, self-supervised contrastive learning (SSCR) \cite{fu2020sscr} was proposed. Underpinning the low recall rate and under construction of objects in GeNeVA-GAN, LatteGAN \cite{matsumori2021lattegan} was proposed recently for better object generation in each step of the iterative manipulation.

%% file: chapters/ch3_methods.tex
\chapter{Methods}
\label{ch3_methods} 

\section{Implementation}
Diffusion model is used to generate artistic design images. Diffusion is based on a concept of non-equilibrium thermodynamics which suggests that all systems in nature are continuously subject to flux of matter or energy. This technique gradually adds noise to an image until it appears to be generated from a normal distribution and then learns to reverse this process by learning to convert noisy image to a less noisy image. At time of sampling the process starts from a pattern of random dots and the learned model gradually converts this pattern into an image guided by the text and reference image. The features of text and reference image were obtained using the contrastive language-image pair (CLIP)\cite{radford2021learning} model which is a representation learner for cross-modal information. 

Manipulated Images were generated using diffusion which refines the noisy image in each step of the reverse process. A U-net \cite{ronneberger2015u} is learned to estimate the amount of noise in the previous step image, also attending to the modification text and reference image in the middle layers of the architecture.

\subsection{Forward Diffusion}

\begin{figure}[h]
  \centering
   \centerline{\includegraphics[width=1\linewidth]{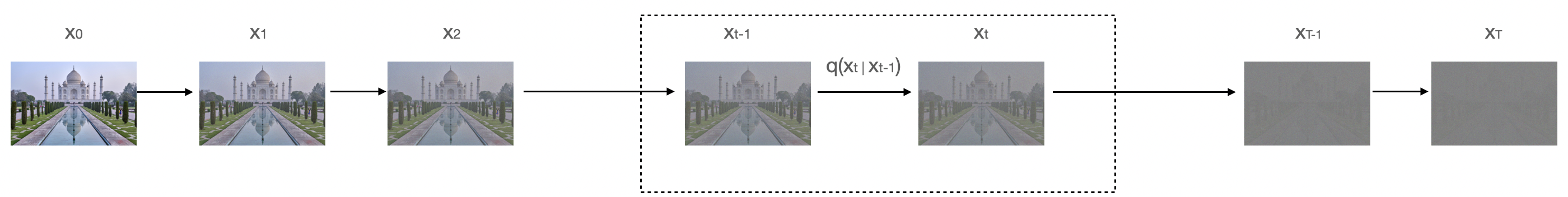}}

   \caption{Forward Diffusion applied on an image \cite{ho2020denoising}}
   \label{fig:for_diff}
\end{figure}

We define a forward noising process(represented using conditional probability distribution $q(x_{t} | x_{t-1})$) which produces latents $x_1$ to $x_T$ by adding Gaussian noise at time $t$ with variance $\beta_{t} \in (0, 1)$ as shown in Fig \ref{fig:for_diff}. What distinguishes diffusion models from other types of latent variable models is that the approximate posterior $q(x_{1:T} | x_{0})$, called the forward process or diffusion process, is fixed to a Markov chain that gradually adds Gaussian noise to the data according to a variance schedule $\beta_{1}$ , . . . , $\beta_{T}$. At every step we assume to generate a noisy image conditioned on the previous image using a normal distribution. This normal distribution, takes the image at the previous step, re-scales it by a factor of $\sqrt(1-\beta_{t})$ and adds a tiny bit of noise with a variance of $\beta_{t}$. The schedule of $\beta$’s is defined such that $\beta_{0}< \beta_{1}< \beta_{2}<…< \beta_{T}$, where T is the last step in forward iteration.

\begin{equation}
    q(x_{t} | x_{t-1}) \sim \mathcal{N}(\sqrt{1-\beta_{t}} x_{t-1},\beta_{t}I)
\end{equation}

We can also define the joint distribution for all the samples that will be generated in this chain of forward diffusion starting from $x_1$ till $x_T$ as

\begin{equation}
    q(x_{1:T} | x_{0}) = \prod_{t=1}^{T} q(x_{t} | x_{t-1})
\end{equation}

Since we are using such a simple distribution to generate samples in forward diffusion, cant we just jump to any $t$'th step in the forward chain using function composition? The answer is, Yes we can by using Normal distribution where mean is actual input re-scaled by $\alpha_{t}$’s and variance is the corresponding noise added to it. Here $\alpha_{t} = 1 - \beta_{t}$ and $\bar\alpha_{t} = \prod_{i=1}^{t} \alpha_{i}$.
\begin{equation}
    q(x_{t} | x_{0}) \sim \mathcal{N}(\sqrt{\bar\alpha_{t}} x_{t-1},(1-\bar\alpha_{t})I)
\end{equation}

\subsection{Reverse Diffusion}

\begin{figure}[h]
  \centering
   \centerline{\includegraphics[width=1\linewidth]{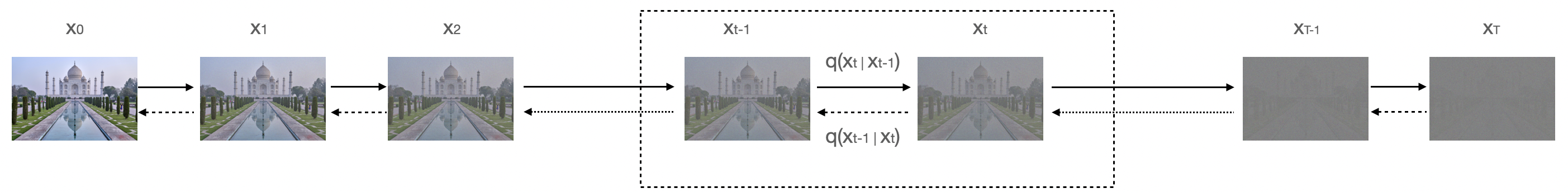}}

   \caption{Reverse Diffusion applied on an image \cite{ho2020denoising}}
   \label{fig:rev_diff}
\end{figure}

So far we talked about forward process of smoothing the data distribution, but now we will focus on defining generative model which reverse the forward diffusion. In order to generate data, we sample a data point from $\mathcal{N}(0, I)$ and transition from noisy latent image to a less noisy latent image using a true denoising distribution as shown in Fig \ref{fig:rev_diff}

Theoretically we can say that $x_{t-1} \sim q(x_{t-1} | x_{t})$ but computing this distribution is intractable. Using Bayes rule we can show that 
\begin{equation}
    q(x_{t-1} | x_{t}) \propto q(x_{t}) * q(x_{t} | x_{t-1})
\end{equation}

where $q(x_{t} | x_{t-1})$ is tractable(diffusion kernel) but marginal $q(x_{t})$ is intractable and so the product is intractable as well. Since we cannot compute it, we try to approximate the required distribution using a normal distribution if the $\beta$ schedule is small in the forward process.

To approximate the reverse distributions, we parameterize the normal distribution using noisy image, which predicts the mean of less noisy image. We can assume a U-net to learn parameters for denoising images. We can also define the joint distribution of full reverse trajectory of the latents.

\begin{equation}
    p_{\theta}(x_{0:T}) = p(x_{T}) \prod_{t=1}^{T} p_{\theta}(x_{t-1} | x_{t})
\end{equation}

where each step of reverse distribution can be approximated as :-
\begin{equation}
     p_{\theta}(x_{t-1} | x_{t}) \sim \mathcal{N}(\mu_{\theta}(x_{t}, t), \Sigma_{\theta}(x_{t}, t))
\end{equation}

\begin{figure}[h]
  \centering
   \centerline{\includegraphics[width=1\linewidth]{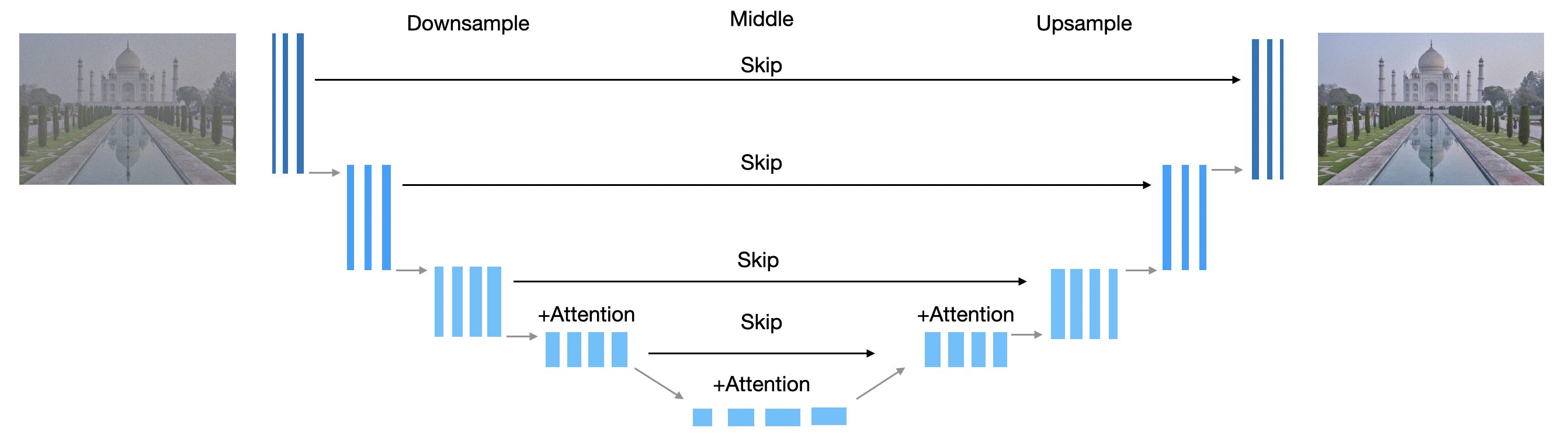}}

   \caption{U-net architecture used to parameterize reverse diffusion process \cite{ho2020denoising}}
   \label{fig:conditionalUnet}
\end{figure}

As represented in Fig \ref{fig:conditionalUnet}, the layers in the encoder part of U-net are skip connected and concatenated with layers in the decoder part. This makes the U-nets use fine-grained details learned in the encoder part to construct an image in the decoder part. These kinds of connections are long skip connections whereas the ones in ResNets \cite{he2016deep} are short skip connections. Main task of Skip connections is to pass earlier-layer semantic information unchanged to the later-layers of the neural architecture.

To guide the design image generation process using description text and rough sketch image, we combine information(CLIP embedding) of both input modalities. These embedding are then passed as the class-conditioning in the diffusion model such that the neural network should generate image with less and less noise but with more guidance from this additional input of the class-conditioning variable obtained from the text, specifying what kind of image to generate.  

We used the attention layer to cross-attend the embedding and middle layers of the U-net architecture. 
This happens at each reverse step. To train the model, we used variational lowerbound objective as proposed in DDPM\cite{ho2020denoising}.

\subsection{Stable Diffusion}
\begin{figure}[h]
  \centering
   \centerline{\includegraphics[width=1\linewidth]{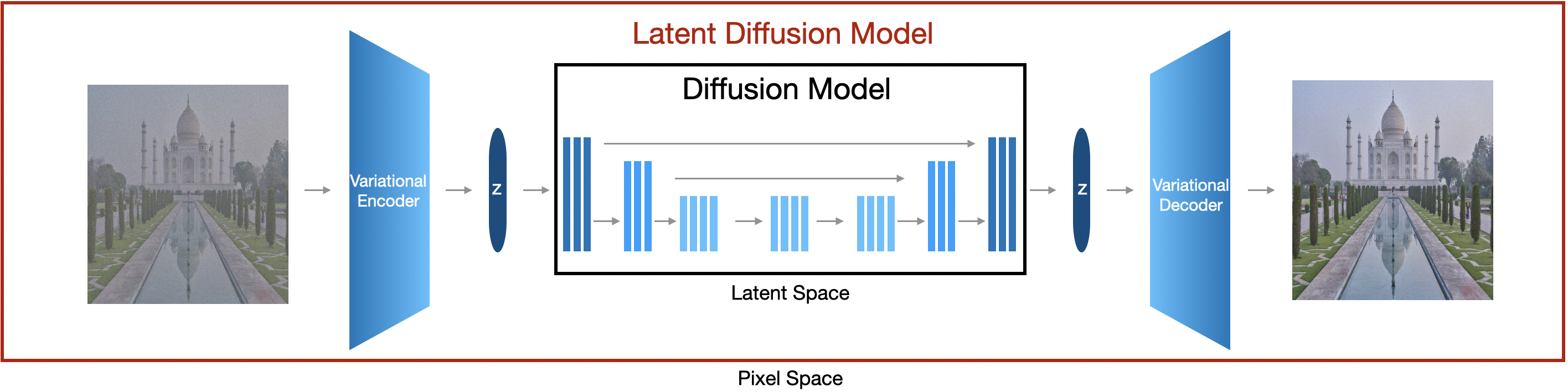}}

   \caption{Architecture of Latent Diffusion model \cite{rombach2022high}}
   \label{fig:stabelDiffusion}
\end{figure}

An issue that remains with diffusion models is that they require a lot of data and compute to train. Because we’re running things for every pixel of an image, and often doing many steps of processing on these pixels, the computation adds up quickly. This computation could be saved if only we had a way to operate on a more compressed representation of an image as produced by Auto-encoders, for example, have a latent space where each point in the latent space maps to an output image. Thus in the latent diffusion model, we first find a perceptually equivalent, but computationally more suitable space, in which we will train diffusion models for high-resolution image synthesis. The overall architecture is shown in fig \ref{fig:stabelDiffusion}.

Latent diffusion model \cite{rombach2022high} convert images to latent space first and then perform the diffusion on latent variables. Therefore the U-net learns parameters suitable for removing noise from latent variable of a high-quality pixel image.

%% file: chapters/ch4_results.tex
\chapter{Results and Discussions}
\label{ch4_results} 

\section{Discussion} 

We provide a way for AI and humans to work collaboratively to produce novel and feasible art designs. Although the human designers are still the driver for final art design, now they could decide which designs generated are good for implementation and refinement purpose in a short time. Industrial applications are utilising principles of generative design to produce unique design patterns that are guided by high level information. This guiding information provides a preconceived notion of the final design. Diffusion models have been proved to be significant for the task of image synthesis to generate images with clear object structures, which could be noted from the images generated by the model. 

\section{Qualitative Analysis}

\begin{figure}[htb]
     \centering
     \begin{subfigure}[b]{0.23\linewidth}
         \centering
         \includegraphics[width=\textwidth]{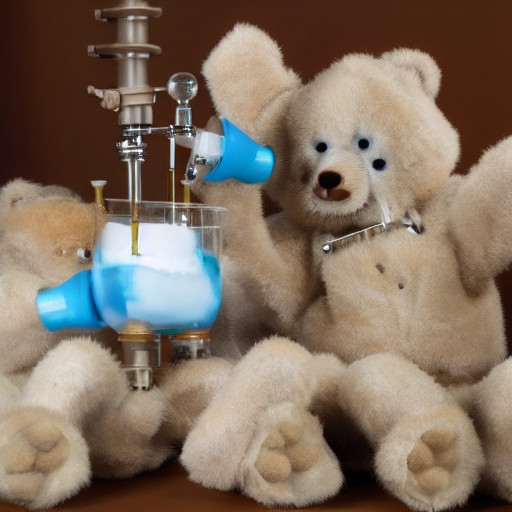}
     \end{subfigure}
     \hfill
     \begin{subfigure}[b]{0.23\linewidth}
         \centering
         \includegraphics[width=\textwidth]{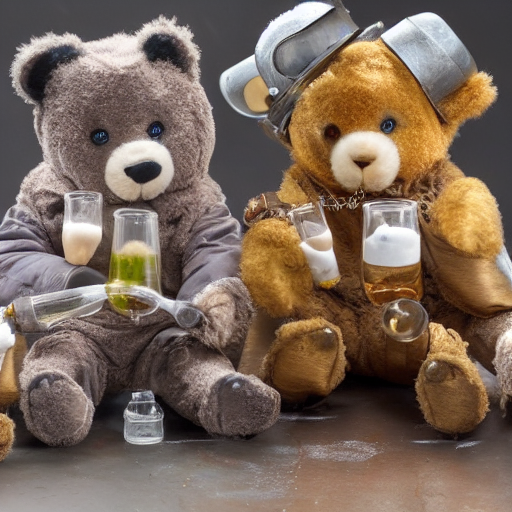}
     \end{subfigure}
     \hfill
     \begin{subfigure}[b]{0.23\linewidth}
         \centering
         \includegraphics[width=\textwidth]{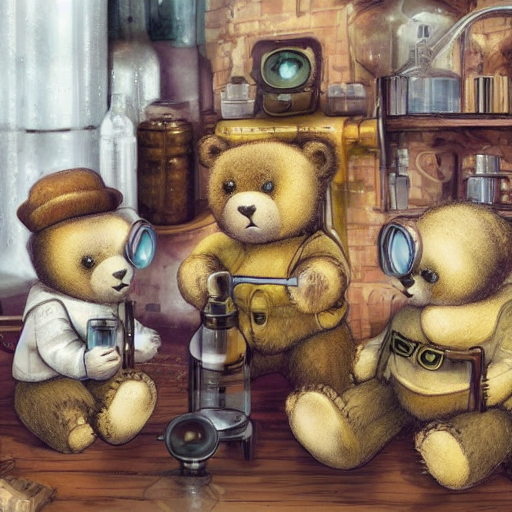}
     \end{subfigure}
     \hfill
     \begin{subfigure}[b]{0.23\linewidth}
         \centering
         \includegraphics[width=\textwidth]{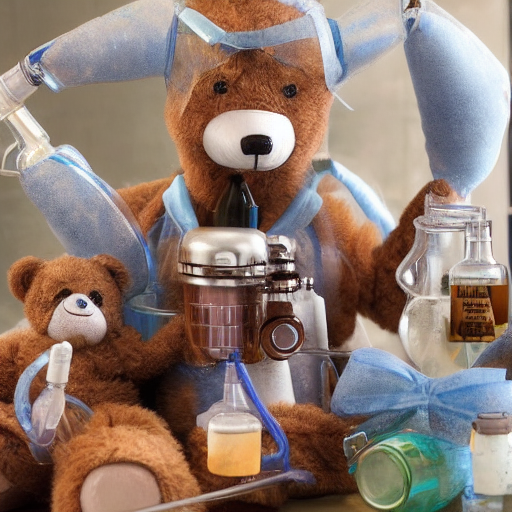}
     \end{subfigure}
        \caption{Generated images for ``Teddy bears mixing sparkling chemicals as mad scientists in a steampunk style"}
        \label{fig:i_1}
\end{figure}

\begin{figure}[htb]
     \centering
     \begin{subfigure}[b]{0.23\linewidth}
         \centering
         \includegraphics[width=\textwidth]{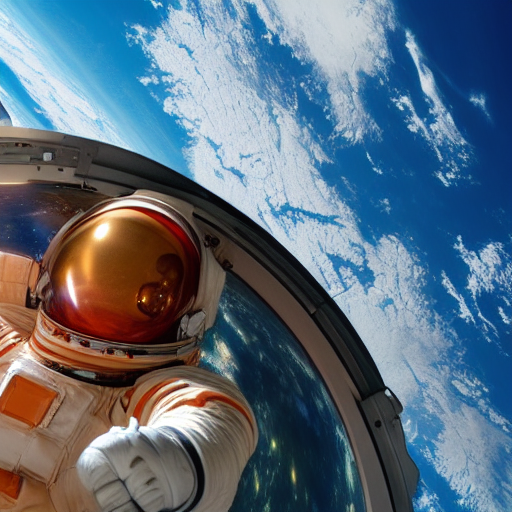}
     \end{subfigure}
     \hfill
     \begin{subfigure}[b]{0.23\linewidth}
         \centering
         \includegraphics[width=\textwidth]{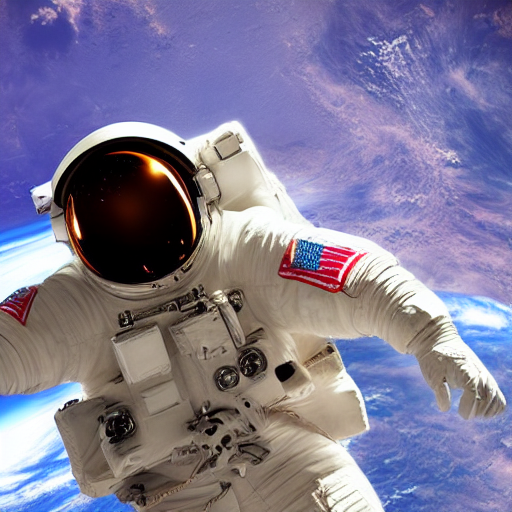}
     \end{subfigure}
     \hfill
     \begin{subfigure}[b]{0.23\linewidth}
         \centering
         \includegraphics[width=\textwidth]{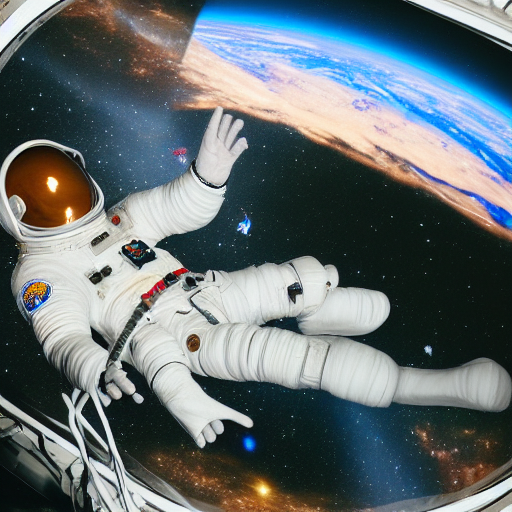}
     \end{subfigure}
     \hfill
     \begin{subfigure}[b]{0.23\linewidth}
         \centering
         \includegraphics[width=\textwidth]{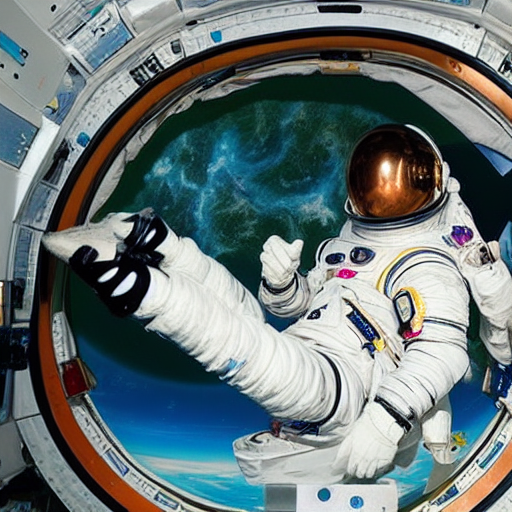}
     \end{subfigure}
        \caption{Generated images for ``An astronaut lounging in a tropical resort in space"}
        \label{fig:i_2}
\end{figure}

\begin{figure}[htb]
     \centering
     \begin{subfigure}[b]{0.23\linewidth}
         \centering
         \includegraphics[width=\textwidth]{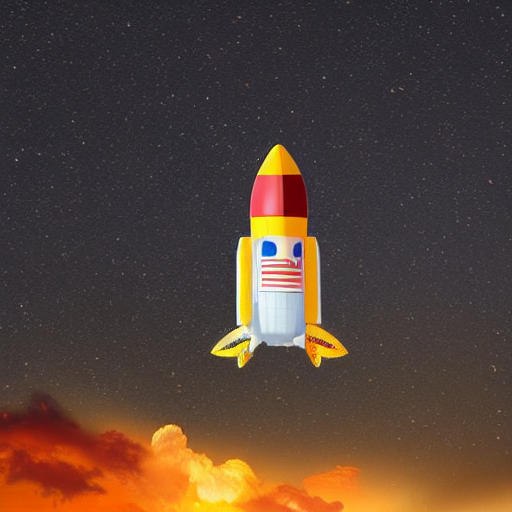}
     \end{subfigure}
     \hfill
     \begin{subfigure}[b]{0.23\linewidth}
         \centering
         \includegraphics[width=\textwidth]{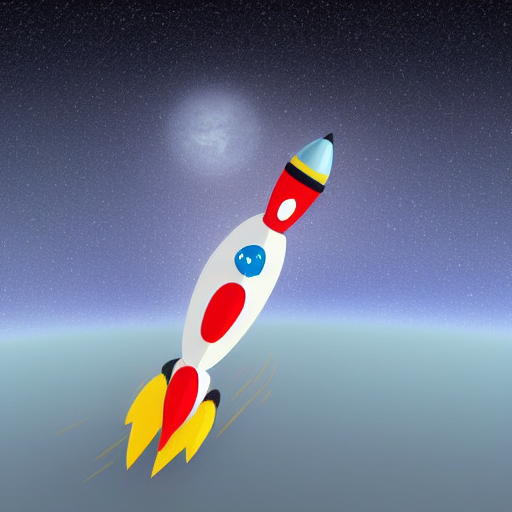}
     \end{subfigure}
     \hfill
     \begin{subfigure}[b]{0.23\linewidth}
         \centering
         \includegraphics[width=\textwidth]{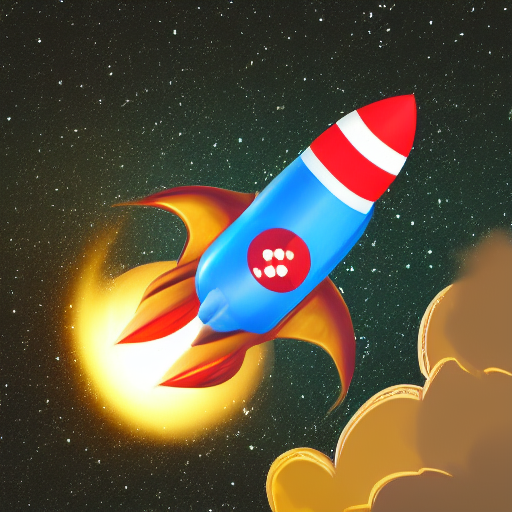}
     \end{subfigure}
     \hfill
     \begin{subfigure}[b]{0.23\linewidth}
         \centering
         \includegraphics[width=\textwidth]{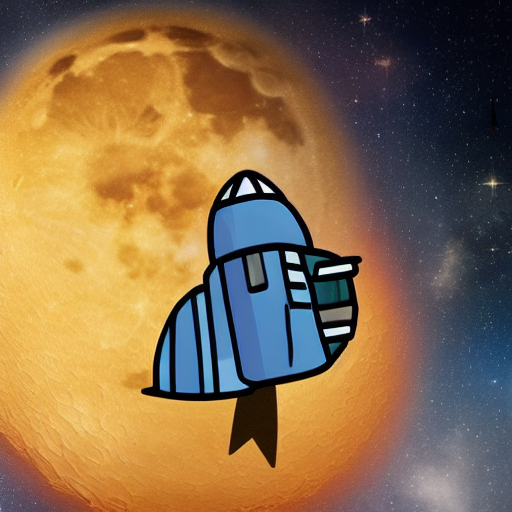}
     \end{subfigure}
        \caption{Generated images for ``A Brain riding rocket ship heading towards moon"}
        \label{fig:i_3}
\end{figure}

\begin{figure}[htb]
     \centering
     \begin{subfigure}[b]{0.23\linewidth}
         \centering
         \includegraphics[width=\textwidth]{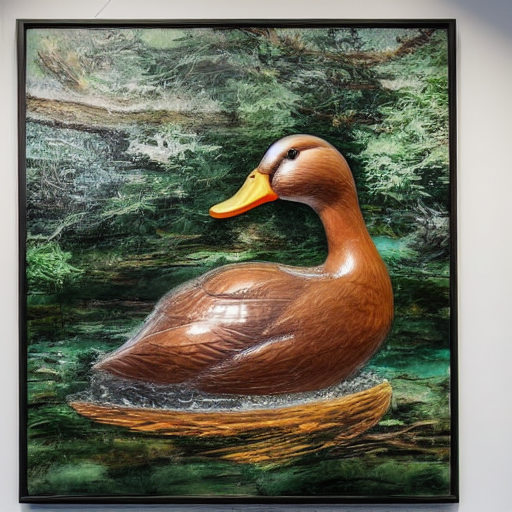}
     \end{subfigure}
     \hfill
     \begin{subfigure}[b]{0.23\linewidth}
         \centering
         \includegraphics[width=\textwidth]{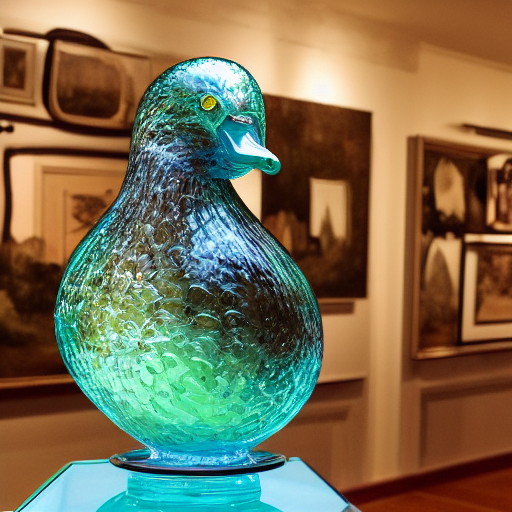}
     \end{subfigure}
     \hfill
     \begin{subfigure}[b]{0.23\linewidth}
         \centering
         \includegraphics[width=\textwidth]{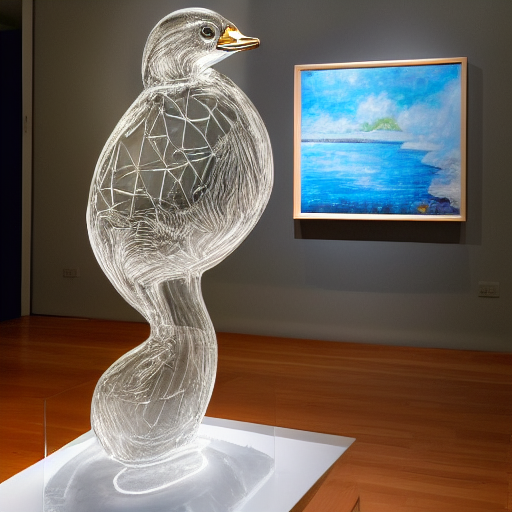}
     \end{subfigure}
     \hfill
     \begin{subfigure}[b]{0.23\linewidth}
         \centering
         \includegraphics[width=\textwidth]{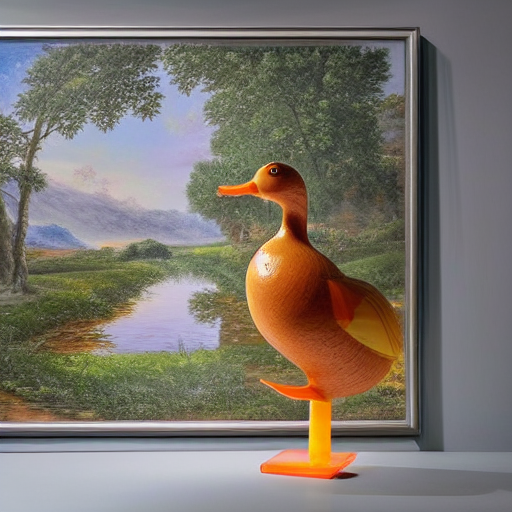}
     \end{subfigure}
        \caption{Generated images for ``A Brain riding rocket ship heading towards moon"}
        \label{fig:i_4}
\end{figure}

\begin{figure}[htb]
    \centering
     \begin{subfigure}[b]{0.23\linewidth}
         \centering
         \includegraphics[width=\textwidth]{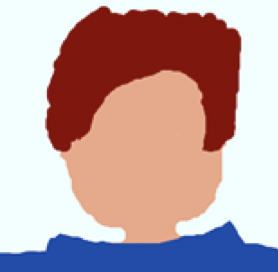}
     \end{subfigure}
     \\\hfill
     \\\hfill
     \centering
     \begin{subfigure}[b]{0.23\linewidth}
         \centering
         \includegraphics[width=\textwidth]{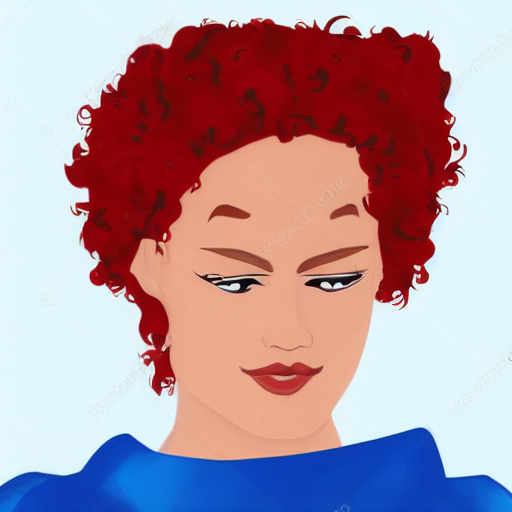}
     \end{subfigure}
     \hfill
     \begin{subfigure}[b]{0.23\linewidth}
         \centering
         \includegraphics[width=\textwidth]{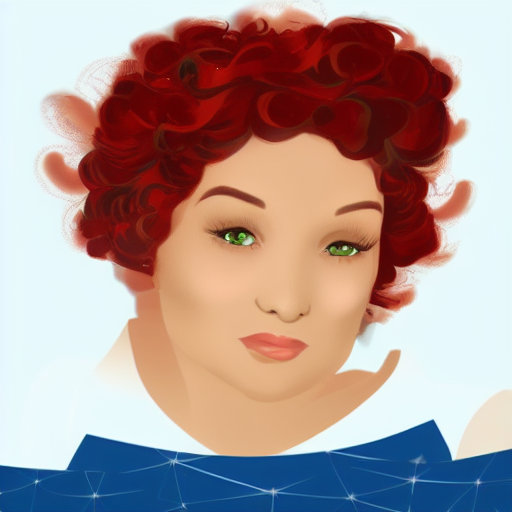}
     \end{subfigure}
     \hfill
     \begin{subfigure}[b]{0.23\linewidth}
         \centering
         \includegraphics[width=\textwidth]{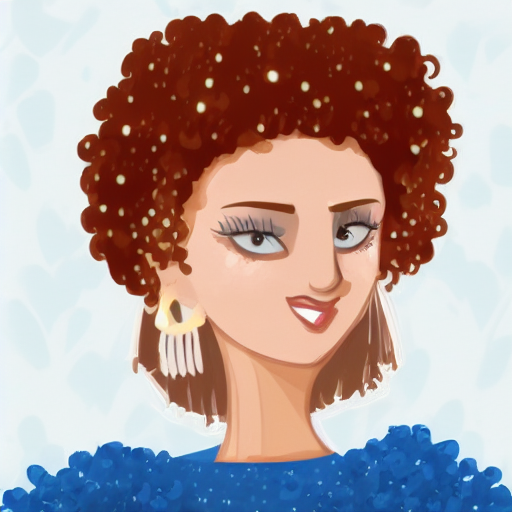}
     \end{subfigure}
     \hfill
     \begin{subfigure}[b]{0.23\linewidth}
         \centering
         \includegraphics[width=\textwidth]{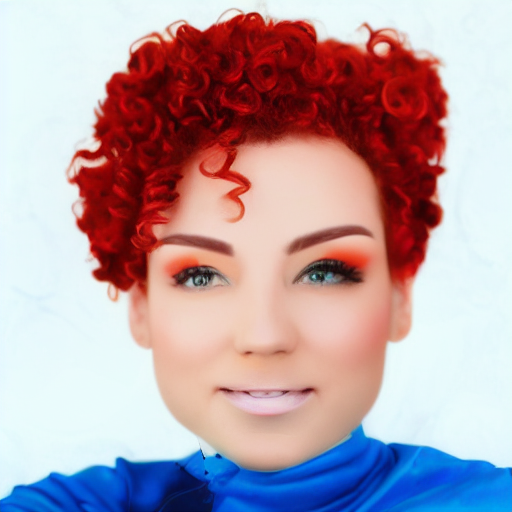}
     \end{subfigure}
        \caption{Generated images for ``Lady with curly hairs and makeup" using top row guiding image and text}
        \label{fig:i_5}
\end{figure}

\begin{figure}[htb]
    \centering
     \begin{subfigure}[b]{0.23\linewidth}
         \centering
         \includegraphics[width=\textwidth]{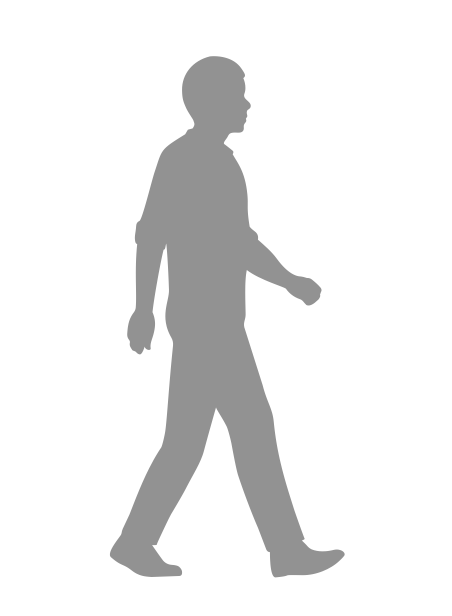}
     \end{subfigure}
     \\\hfill
     \\\hfill
     \centering
     \begin{subfigure}[b]{0.23\linewidth}
         \centering
         \includegraphics[width=\textwidth]{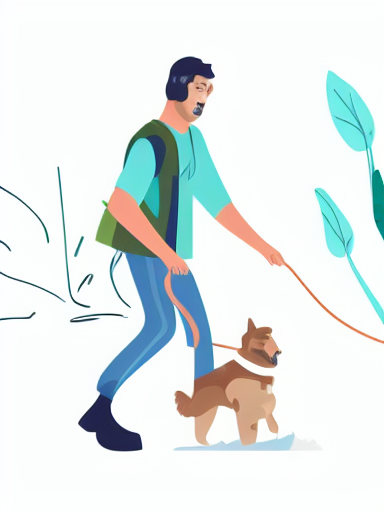}
     \end{subfigure}
     \hfill
     \begin{subfigure}[b]{0.23\linewidth}
         \centering
         \includegraphics[width=\textwidth]{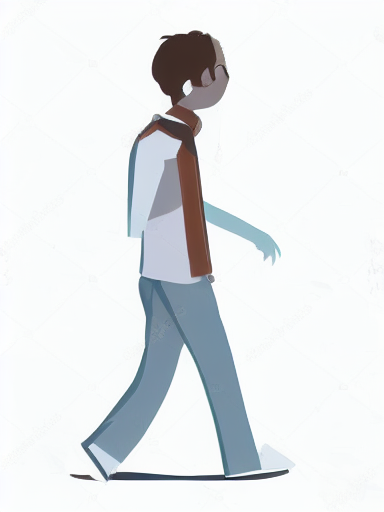}
     \end{subfigure}
     \hfill
     \begin{subfigure}[b]{0.23\linewidth}
         \centering
         \includegraphics[width=\textwidth]{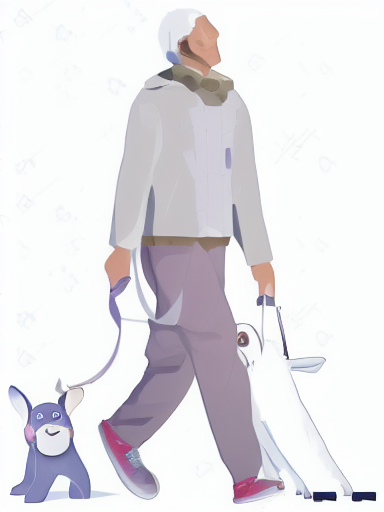}
     \end{subfigure}
     \hfill
     \begin{subfigure}[b]{0.23\linewidth}
         \centering
         \includegraphics[width=\textwidth]{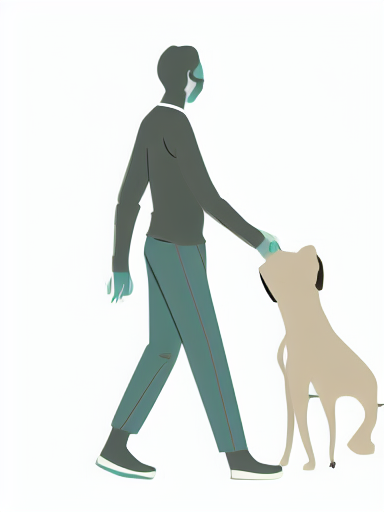}
     \end{subfigure}
        \caption{Generated images for ``Man walking in a park with his dog" using top row guiding image and text}
        \label{fig:i_6}
\end{figure}

\begin{figure}[htb]
    \centering
     \begin{subfigure}[b]{0.23\linewidth}
         \centering
         \includegraphics[width=\textwidth]{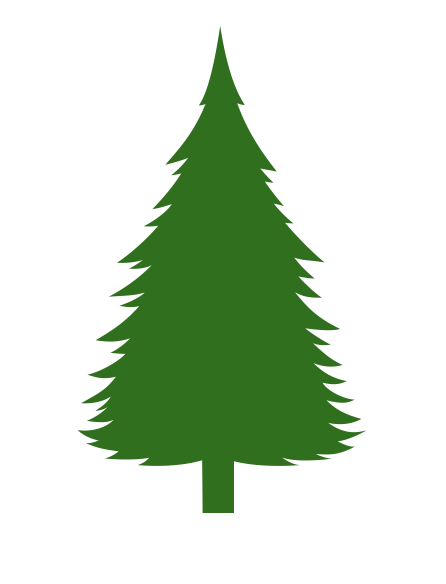}
     \end{subfigure}
     \\\hfill
     \\\hfill
     \centering
     \begin{subfigure}[b]{0.23\linewidth}
         \centering
         \includegraphics[width=\textwidth]{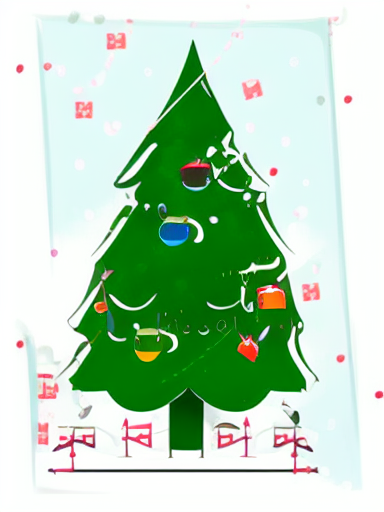}
     \end{subfigure}
     \hfill
     \begin{subfigure}[b]{0.23\linewidth}
         \centering
         \includegraphics[width=\textwidth]{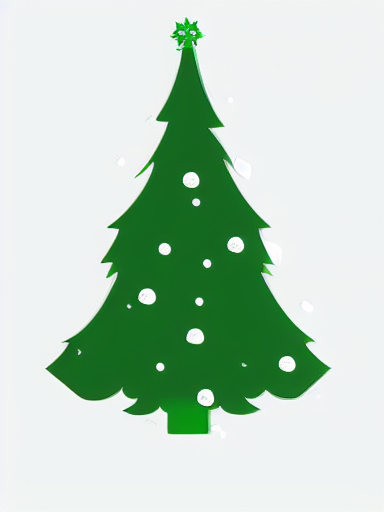}
     \end{subfigure}
     \hfill
     \begin{subfigure}[b]{0.23\linewidth}
         \centering
         \includegraphics[width=\textwidth]{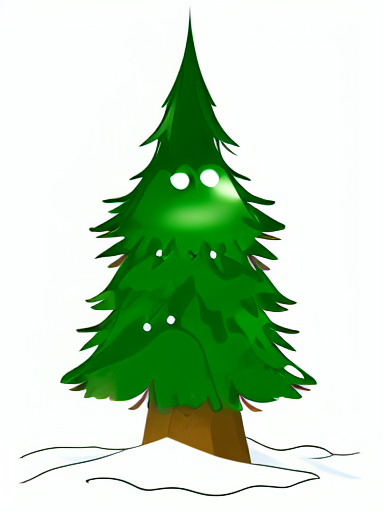}
     \end{subfigure}
     \hfill
     \begin{subfigure}[b]{0.23\linewidth}
         \centering
         \includegraphics[width=\textwidth]{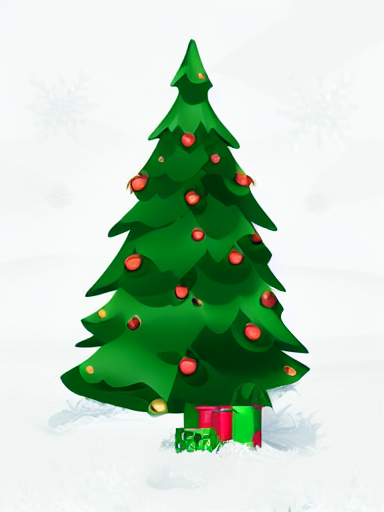}
     \end{subfigure}
        \caption{Generated images for ``Tree covered with snow in a winter Christmas season" using top row guiding image and text}
        \label{fig:i_7}
\end{figure}

The qualitative results of art generated using architecture mentioned in Chapter \ref{ch3_methods} are shown below. Each table contain four columns listing the text used to guide the product generation process, followed by generated images. Then the rough sketch for product generation is shown along with text-and-sketch guided product designs. Art designs were generated for many text prompts out of which some are shown ahead. 
\begin{enumerate}
    \item Art designs for ``Teddy bears mixing sparkling chemicals as mad scientists in a steampunk style" \cite{ramesh2021zero} in figure \ref{fig:i_1}
    \item Art designs for ``An astronaut lounging in a tropical resort in space" \cite{ramesh2021zero} in figure \ref{fig:i_2}
    \item Art designs for ``A Brain riding rocket ship heading towards moon" \cite{saharia2022photorealistic} in figure \ref{fig:i_3}
    \item Art designs for ``A transparent sculpture of a duck made out of glass. The sculpture is in front of a painting of a landscape" \cite{saharia2022photorealistic} in figure \ref{fig:i_4}
    \item Rough sketch image and text guided art designs for ``Lady with curly hairs and makeup" in figure \ref{fig:i_5}
    \item Rough sketch image and text guided art designs for ``Man walking in a park with his dog" in figure \ref{fig:i_6}
    \item Rough sketch image and text guided art designs for ``Tree covered with snow in a winter Christmas season" in figure \ref{fig:i_7}
\end{enumerate}

Goal of this project was to create art design utilising the high level direction provided by the designer using text or image modality. The proposed method was found useful to generate novel designs using latent space diffusion model which take less time and compute compared to diffusion technique in pixel space. 

The illustrative examples shown in figures \ref{fig:i_1} till \ref{fig:i_4} generated using only text are astonishing. Images, such as illustrations, paintings, and photographs, can often be easily described using text, but can require specialized skills and hours of labor to create. Therefore, a tool capable of generating realistic images from natural language can empower humans to create rich and diverse visual content with unprecedented ease. The ability to edit images using natural language further allows for iterative refinement and fine-grained control, both of which are critical for real world applications. The photorealism of arts generated in figure \ref{fig:i_4} is astonishing. You could note that the image generated from text is aware of the basic concept of physics  as reflection and transparent nature of objects formed from glass through which colors of anything lying at the back could be seen and therefore the wall/background colors are visible through the glass. Talking of rough sketch effect, given only sketch and guiding text in figures \ref{fig:i_5}, \ref{fig:i_6} and \ref{fig:i_7}, the generation results were highly affected by input distribution.

But one must be wondering about the need to use diffusion techniques when others like auto-regressive models and GANs already exist. Well, ADM paper \cite{dhariwal2021diffusion} already proved that Diffusion models can beat GAN on image generation from the concern of image fidelity and photo-realism.  An intuitive reason is that a GAN starts from a random noise and is expected to produce an image which then a discriminator accepts or rejects, but a diffusion model is much slower, iterative and guided process. In the reverse process, there is very little room for going very far astray and in each step of backward diffusion add more and more details to the random noise. This assures that the diffusion model is more faithful than GANs for the during the sampling process. While autoregressive models are data hungry and used a lot of parameters to train. 

%% file: chapters/ch5_conclusion.tex
\chapter{Conclusions and Future Scope}
\label{ch5_conclusion} 

The results mentioned in figure \ref{fig:i_1}, \ref{fig:i_2}, \ref{fig:i_3}, \ref{fig:i_4}, \ref{fig:i_5}, \ref{fig:i_6} and \ref{fig:i_7} are interesting to note the effect of providing guiding modality data in each scenario. While the text helped in explaining objects expected in generated designs, the rough sketch images helped to provide more structure during generation. 

These findings suggest that diffusion models are suitable for quality art generation tasks even with single or multiple guidance signals. 

\section{Impact of research}

Diffusion technique open doors to a new paradigm of learning for generative modelling apart from existing likelihood based models such as auto-regressive  models, variational autoencoders and adversarial networks. 

Industries could use this technique for other generative purposes such as language generation in sequence to sequence learning tasks or image generation in higher dimension. Through this project we realised the power of diffusion models for quality image generation tasks, by providing the guidance through too many external signals such as text and rough sketch images. 

Diffusion modelling code developed during this project could be utilised for other tasks such as image-to-image manipulation, text-to-image generation, image colorisation, image inpainting, open domain text based question answering, language translation, visual question answering, text paraphrasing, 3D object generation and other similar tasks which involve conditional or unconditional generation of some modality of data. 

We conclude that diffusion technique is helpful as a product design generator that could leverage the guidance provided by different modality of input such as either text or image.

\section{Limitations and Future work}

Although the quality of images generated using diffusion in latent space is helpful to produce novel product designs, but this technique has some social biases that it learned during the training process. 

\begin{figure}[htb]
     \centering
     \begin{subfigure}[b]{0.23\linewidth}
         \centering
         \includegraphics[width=\textwidth]{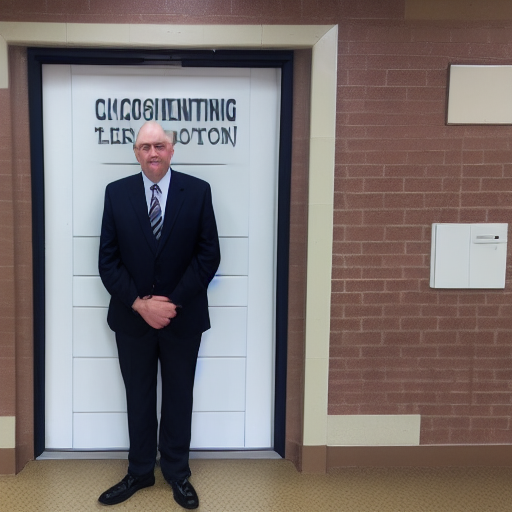}
     \end{subfigure}
     \hfill
     \begin{subfigure}[b]{0.23\linewidth}
         \centering
         \includegraphics[width=\textwidth]{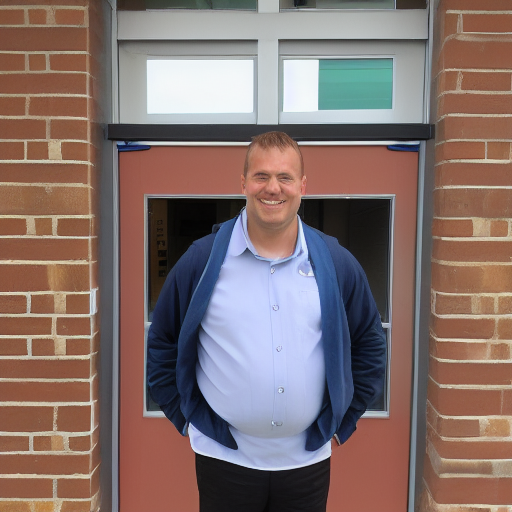}
     \end{subfigure}
     \hfill
     \begin{subfigure}[b]{0.23\linewidth}
         \centering
         \includegraphics[width=\textwidth]{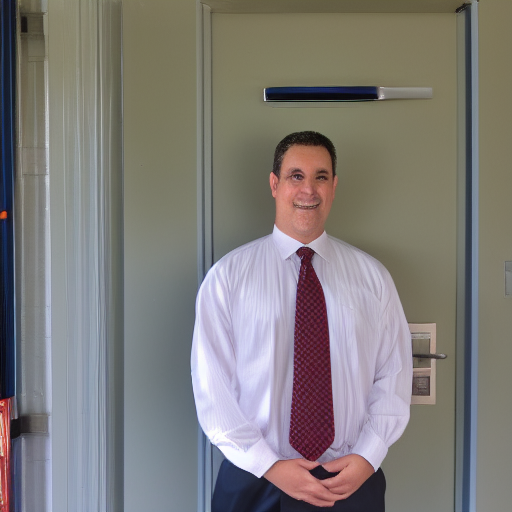}
     \end{subfigure}
     \hfill
     \begin{subfigure}[b]{0.23\linewidth}
         \centering
         \includegraphics[width=\textwidth]{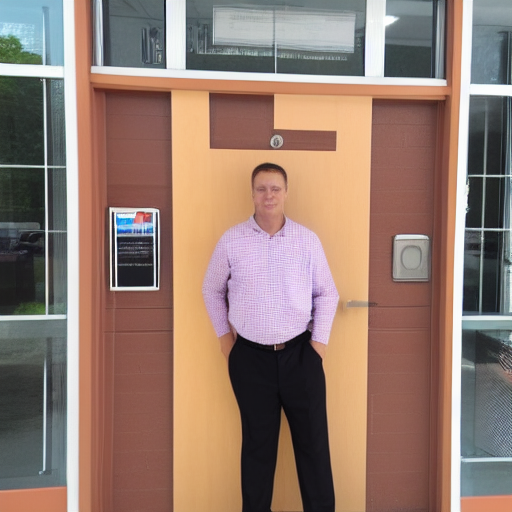}
     \end{subfigure}
        \caption{Generated images for ``Principal standing in front of the school door"}
        \label{fig:limit_1}
\end{figure}

\begin{figure}[htb]
     \centering
     \begin{subfigure}[b]{0.23\linewidth}
         \centering
         \includegraphics[width=\textwidth]{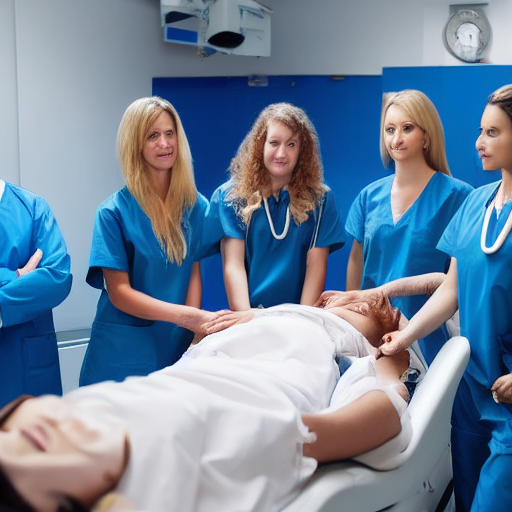}
     \end{subfigure}
     \hfill
     \begin{subfigure}[b]{0.23\linewidth}
         \centering
         \includegraphics[width=\textwidth]{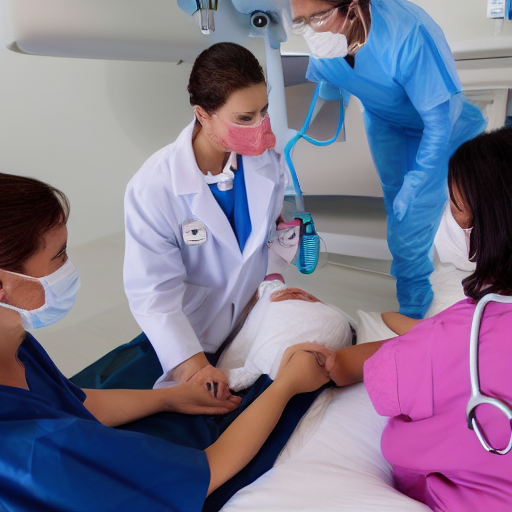}
     \end{subfigure}
     \hfill
     \begin{subfigure}[b]{0.23\linewidth}
         \centering
         \includegraphics[width=\textwidth]{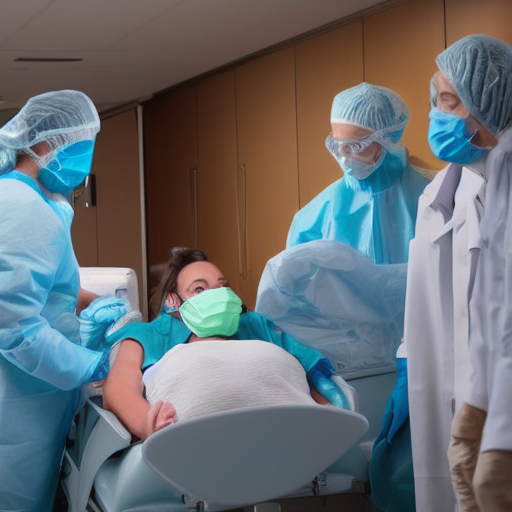}
     \end{subfigure}
     \hfill
     \begin{subfigure}[b]{0.23\linewidth}
         \centering
         \includegraphics[width=\textwidth]{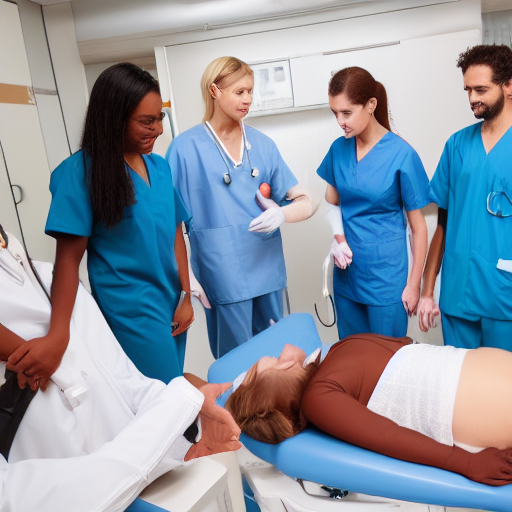}
     \end{subfigure}
        \caption{Generated images for ``Nursing staff helping doctor in the operation theater"}
        \label{fig:limit_2}
\end{figure}

For illustration, the images obtained using the text ``Principal standing in front of the school door'' are shown in fig \ref{fig:limit_1}. While the generation results for text ``Nursing staff helping doctor in the operation theater" are shown in fig \ref{fig:limit_2}. In the former case, all the principal characters are male while in later, all the nursing staff is mostly depicted by feminine characters. 

There is also a risk of copyright infringement, as the model might be inspired by art from existing artists as they are trained on images from the web. So it becomes important in the future extension of this work to tackle these problems.